\documentclass[11pt,a4paper]{article}

\usepackage[utf8]{inputenc}
\usepackage[T1]{fontenc}
\usepackage[margin=1in]{geometry}
\usepackage{amsmath,amssymb,amsthm}
\usepackage{graphicx}
\usepackage{tikz}
\usetikzlibrary{arrows.meta, positioning, shapes, calc, decorations.pathreplacing, fit, backgrounds}
\usepackage{caption}
\usepackage{subcaption}
\usepackage{authblk}
\usepackage{booktabs}
\usepackage{hyperref}
\hypersetup{
  colorlinks=true,
  linkcolor=blue!50!black,
  citecolor=blue!50!black,
  urlcolor=blue!60!black
}
\usepackage[square,numbers]{natbib}

\usepackage{titlesec}
\titleformat{\section}{\normalfont\Large\bfseries}{\thesection}{1em}{}
\titleformat{\subsection}{\normalfont\large\bfseries}{\thesubsection}{1em}{}

\title{\textbf{How to Build Marcus's Algebraic Mind:
Algebro-Deterministic Substrate over Galois Fields}}

\author[1]{Hiroyuki Chuma\thanks{Corresponding author: \texttt{chuma@iir.hit-u.ac.jp}}}
\author[2]{Kanji Otsuka}
\author[3]{Yoichi Sato}
\affil[1]{Institute of Innovation Research, Hitotsubashi University, Kunitachi, Tokyo 186-8603, Japan (Professor Emeritus)}
\affil[2]{Meisei University, Hino, Tokyo, Japan (Professor Emeritus)}
\affil[3]{Shuhari System, Tokyo, Japan}

\date{\today}

\begin{document}

\maketitle
\tableofcontents

\begin{abstract}
\noindent This paper reports a correspondence that neither of its two sides was designed to produce. In \emph{The Algebraic Mind}, \citet{Marcus2001} identified three components that any adequate cognitive architecture must support---operations over variables, recursively structured representations, and a distinction between mental representations of individuals and of kinds---and argued that the multilayer perceptron, as standardly used, supports none of them. He closed with the explicit acknowledgment that a neural implementation of these components, organized around registers and \emph{treelets} and constructed by cascading developmental programs rather than by gradient descent, remained a programmatic conjecture. Twenty-five years later a memory architecture built for entirely unrelated reasons---speed, power, and cost on commodity silicon---turns out to satisfy that specification operation for operation, and to satisfy it not through three separate mechanisms but through a single elementary one.

PyVaCoAl/VaCoAl \cite{Chuma2026a} is a hyperdimensional computing (HDC) architecture organized end-to-end around one algebraic primitive: XOR-and-shift over GF(2), realised by primitive-polynomial linear-feedback shift registers (LFSRs). It supports reversible variable binding through $\mathrm{Bind}(R,F) = R \oplus \mathrm{shift}(F)$, non-commutative compositional bundling that distinguishes \emph{the dog bites the man} from \emph{the man bites the dog}, and address-space individual/kind separation, all under the same algebra and at fixed dimension.

We make three scoped claims. \textbf{Capability}: exact reversible binding over GF(2) at $O(L)$ cost supplies, as primitives of the architecture rather than as emergent properties of training, each of Marcus's three pillars, together with the post-hoc inspectability that his Pillar 2 most wanted and that no lossy or learned substrate offers. \textbf{Necessity}: that the three pillars are supported by one operation is not a convenience of exposition but a constraint. Weaken the elementary operation to an approximate inverse---as circular convolution does---and all three degrade together, because a representation that cannot be exactly taken apart was never composed in the first place; it was mixed. Compositionality is purchased with decomposability, and the purchase is made at the level of the elementary operation or not at all. \textbf{Position}: we do not claim that this architecture is the mind, that the brain is an LFSR, or that we surpass large language models. The substrate is orthogonal to them: it supplies the reversible, auditable, symbol-manipulating layer that statistical minds structurally lack, and the natural division of labour is linguistic fluidity on one side and compositional integrity on the other.

We develop the correspondence pillar by pillar, reinterpret the treelet as an algebraic register set whose identity is a primitive generator polynomial $G_b(x)$, summarise the companion Perspective's reading of the dentate gyrus--CA3 circuit as a biological instance of the same engine \citep{Chuma2026b}, and argue that the engine addresses the specific failure modes that defeated tensor products, holographic reduced representations, and temporal synchrony in 2001. We also show that Marcus's diagnostic demand---that an inflection rule extend to a pseudoverb never encountered---is a rung-3 query in Pearl's sense, and that free generalization and counterfactual reasoning are therefore the same structural capacity under two descriptions.

Claims about biological realisation are stated as structural correspondence throughout. Bit-exact reversibility holds in silicon and only approximately under biological noise; no cognitive experiment is reported here; and the correspondence between developmental programs and primitive-polynomial selection is structural rather than proven by any functional-equivalence theorem.

\vspace{0.5em}
\end{abstract}

\section{Introduction}

\subsection{The 2001 question}

\citet{Marcus2001} closed \emph{The Algebraic Mind} by stating, with characteristic directness, what remained to be done. Even if the components of symbol manipulation---symbols, relations between variables, structured representations, and representations for individuals---turn out to be genuine components of cognition, we still need to discover how those components are implemented in neural hardware. Marcus had a candidate, sketched in chapter 4.4 and chapter 6.3 of the book: registers, implemented as inter- or intracellular circuits, hierarchically arranged into structures he called \emph{treelets}; and a developmental mechanism in which cascades of master control genes, rather than learning, would build the relevant microcircuitry. He was explicit that the proposal was plausible but speculative, with the precise mechanisms of axon guidance, the algebra of registers, and the empirical validation of the developmental program left as targets for future work.

Twenty-five years after \emph{The Algebraic Mind} appeared, two of these targets have been substantially advanced. On the biological side, dendritic XOR-like computation has been directly demonstrated in human cortical pyramidal cells \citep{Gidon2020,Benavides2020}, mossy-fiber detonator transmission has been quantified \emph{in vivo} \citep{Henze2002,Vyleta2016,Chamberland2018}, and developmentally programmed mossy-fiber target specificity has been established at single-cell resolution \citep{Galimberti2006,Rollenhagen2010,Wilke2013}. On the computational side, hyperdimensional computing (HDC) has matured from the holographic reduced representations of \citet{Plate1995} into a family of substrate-aware architectures, the most recent of which abandons the lossy quasi-inverse of circular convolution in favor of bit-exact algebraic operations over GF(2).

What has not been advanced, until now, is the joint. The biology supplies mechanisms without an algebra; the HDC literature supplies algebras without a commitment to any particular tissue. Marcus's own proposal names the missing piece precisely and declines to fill it in: the treelet is a proposal about how hierarchical structures can be arranged in registers, and explicitly \emph{not} a proposal about what operations the registers support. This paper takes the position that the missing operation is available, that it is a single one, and that having it changes the status of the 2001 program from ``symbol manipulation must be possible in neural tissue'' to ``here is an algebra under which it is possible, and here is what the substrate would have to look like.''

\subsection{The Engineering Barrier of ``Approximate Binding''}

It is worth being explicit about why the joint has stayed open, because the obstacle is not a shortage of proposals. Every scheme Marcus reviewed in Chapter 4 supplies binding; what none of them supplies is binding that is simultaneously reversible, non-commutative, and affordable at fixed dimension. Each purchases one property at the price of another, and it is the trade rather than any single failure that has kept the program programmatic.

Tensor products \citep{Smolensky1990} buy exact unbinding and pay in dimension: the representation of a $k$-level recursive structure has dimension equal to the product of $k$ constituent dimensions, and Marcus computes the concrete bill---on the order of $10^7$ nodes for a five-level embedding at reasonable per-role precision (Chapter 4.3.3). Circular convolution \citep{Plate1995} buys fixed dimension and pays in exactness: unbinding returns the filler plus a noise term drawn from the same algebra as the signal, so crosstalk accumulates with bundle size and bounds recursion depth in practice. Temporal synchrony \citep{Shastri1993} and asynchrony \citep{Love1999} buy both and pay in timing precision: discrimination among bindings is linear in the available temporal resolution, and Marcus develops in Chapter 4.3.4 the worry that the required resolution outruns what biological circuits can hold.

Learned dense embeddings, the dominant contemporary approach, do not appear on this list because they decline the trade rather than making it. They provide no elementary binding operation at all; structure is expected to emerge from gradient descent on relational prediction. Marcus's Chapter 3 critique applies to them without modification, and the intervening twenty-five years of scale have not touched it: an architecture that has no operation defined over variables does not acquire one by being given more parameters.

The missing link, then, is an elementary operation that is reversible \emph{and} non-commutative \emph{and} uniform over values, at a fixed and modest cost. Section 3 argues that GF(2) supplies exactly one such operation, and that having only one is the point.

\subsection{The Core of This Study}

\begin{table}[t]
\centering
\caption{Comparison of the 2001 binding alternatives and the VaCoAl engine. The right column is not a claim of superiority on every axis; it is a statement of which trade the engine declines to make.}
\label{tab:axes}
\begin{tabular}{p{3.5cm}p{5.6cm}p{5.3cm}}
\toprule
\textbf{Comparison axis} & \textbf{2001 alternatives} & \textbf{VaCoAl} \\
\midrule
Binding operation & Outer product; circular convolution; spike timing & XOR-and-shift over GF(2) \\
\addlinespace
Unbinding & Contraction (exact); quasi-inverse (lossy); re-synchronisation (timing-bound) & Exact algebraic inverse \\
\addlinespace
Role asymmetry & Tensor order; permutation; firing order & Shift permutation, structural \\
\addlinespace
Cost of recursion & Multiplicative growth in dimension; fixed dimension with accumulating crosstalk; increasing temporal precision & Fixed dimension; bounded by block collision \\
\addlinespace
Inspectability & Exact but unaffordable; approximate; timing-dependent & Bit-exact, auditable post hoc \\
\addlinespace
Reported failure & Dimensional explosion; crosstalk accumulation; precision wall & Block collision, reported as CR1/CR2 \\
\bottomrule
\end{tabular}
\end{table}

The proposal of this paper is that the substrate Marcus's program called for is an algebra rather than a mechanism, and that the algebra is the one Galois bequeathed. PyVaCoAl/VaCoAl replaces probabilistic projection and analog mimicry alike with strictly deterministic digital logic---diffusion over GF(2) by primitive-polynomial LFSRs---and then declines to add anything else. Every operation in the architecture, from address generation to compositional bundling to unbinding, decomposes into XOR and shift. Table~\ref{tab:axes} summarises the resulting structural differences from the 2001 alternatives.

The consequence that concerns us is not the efficiency but the parsimony. Marcus's three pillars are usually treated as three requirements, to be met by three mechanisms, and every architecture we know of that meets more than one meets them by addition. The engine meets all three by subtraction: there is one elementary operation, and the pillars are what it looks like when queried from three directions. Section 10 argues that this is not an accident of exposition but a structural constraint, and that the constraint runs in the unwelcome direction as well---that an architecture which weakens the elementary operation loses all three pillars at once, and cannot repair any of them at a higher level.

\subsection{What Is and Is Not Claimed}

Because this paper joins an engineering substrate to a twenty-five-year-old cognitive specification, it is worth stating at the outset which claims are load-bearing and which are not.

\paragraph{Capability.}
Exact reversible binding over GF(2), $\mathrm{Bind}(R,F) = R \oplus \mathrm{shift}(F)$, computed at $O(L)$ cost and inverted exactly, supplies each of Marcus's three pillars as an architectural primitive rather than as an emergent property of training. Free generalization to an unseen filler requires no retraining because the operation is defined by the role and not by the filler. Role-asymmetric structures are distinguishable because shift acts asymmetrically on the arguments. Individuals and kinds are separable because they are addressed by distinct queries in a shared space. We claim that the resulting bit-exact compositional inspection is structurally unavailable to lossy or learned substrates, which can report a nearest neighbour but not a constituent.

\paragraph{Necessity.}
We claim that the three pillars and the exactness of the elementary operation are the same phenomenon, not four independent features. Replace the exact inverse with a quasi-inverse and every recovered filler arrives carrying a noise term in the same algebra as the signal; at that point the recovery is a similarity judgment rather than a decomposition, the role-asymmetry test degrades to a threshold comparison, and the individual/kind separation collapses into whatever the noise floor permits. A representation that cannot be exactly taken apart was never composed; it was mixed. This is developed in Section 10, and it cuts against the reflex of treating approximation as a tunable quality dial: here it is a change of type.

\paragraph{Position.}
We do not claim that PyVaCoAl/VaCoAl is the mind, that the brain is literally an LFSR, or that we surpass large language models, and we run no comparison against them on any language task. The claim is narrower and, we think, more useful: the architectural commitments of the engine correspond, component for component, to what Marcus identified in 2001 as the missing pieces of a neurally plausible symbol-manipulating substrate, and the correspondence is precise enough to be testable. The natural division of labour is linguistic fluidity on one side and compositional integrity on the other.

\paragraph{Not claimed.}
No cognitive experiment is reported in this paper. Marcus's Chapter 3 inflection studies and Chapter 5 individuation studies have not been re-run on the architecture; doing so is a concrete test, stated as such in Section 13.2, not a result. Bit-exact reversibility holds in silicon and only approximately under biological noise. The biological reading of Section 8 is summarised from the companion Perspective \citep{Chuma2026b} and is defended there, not here. The correspondence between developmental programs and the selection of a primitive generator $G_b(x)$ is structural; no functional-equivalence theorem is offered. And the treelet reinterpretation of Section 7 is one reading of Marcus's proposal, argued from parsimony rather than from necessity.

\subsection{Why Free Generalization Is a Counterfactual Quantity}

Marcus's diagnostic for Pillar 1 is a pseudoverb. Told that the past tense of \emph{walk} is \emph{walked}, a competent speaker will produce \emph{blicketed} for \emph{blicket} without hesitation and without ever having encountered the word. The demand looks like a demand for coverage, and it is easy to read the failure of trained networks to meet it as a shortage of data. It is worth being explicit about why it is not, because the connection to a different literature is easy to mistake for a metaphor.

Asking for the past tense of \emph{blicket} is asking what the system would output along a road it has never travelled. A function approximator can report only where it has been: its outputs are a record of the input region the training distribution actually covered, and its behaviour outside that region is whatever the interpolation happens to do. Enlarging the corpus adds examples; it does not add an operation. The difference is one of type, not of size---the same distinction, in a different vocabulary, that \citet{Pearl2009,Pearl2018} draws between a rung-1 quantity estimated from observation and a rung-2 or rung-3 quantity that requires a mechanism to evaluate. \emph{What would the output be if the variable held a value it has never held} is not a question about the observed distribution at all. Marcus called the required object a universally quantified one-to-one mapping; Pearl would call the query a counterfactual. They are the same demand.

Two structural properties are required to satisfy it, and the engine supplies both. \emph{Uniformity}: because $\mathrm{Bind}(R, F)$ is a fixed function of $F$ parameterized only by $R$, substituting an unseen filler is not an extrapolation but an application---the operation was never defined in terms of the fillers it had seen. \emph{Reversibility}: because the bind is exactly invertible, one can unbind at a role, inspect what occupies it, substitute a different filler, and re-bind, without disturbing any other binding in the structure. That triple---unbind, substitute, rebind---is the operational content of Pearl's $\mathrm{do}(\cdot)$, and Section 12 develops the point. For the moment the narrower observation is enough: Marcus's Pillar 1 and Pearl's ladder are not two programs that happen to be compatible. They are one capacity described twice, and an architecture that supplies it for inflection supplies it for intervention.

\subsection{Structure of the Paper}

Section 2 reviews Marcus's three pillars and states the three open questions his program left for neural implementation. Section 3 introduces the substrate at the level of operational commitments and argues that GF(2) forces the choice of elementary operation rather than merely permitting it. Sections 4--6 develop, pillar by pillar, the correspondence between Marcus's specifications and the engine's commitments. Section 7 reinterprets the treelet as an algebraic register set indexed by a primitive generator polynomial. Section 8 summarises the companion Perspective's reading of DG--CA3 as a biological instance of the same engine, and states carefully what kind of equivalence is being claimed. Section 9 compares the result with the 2001 alternatives.

Section 10 is the core of the paper: it isolates the general form of the necessity claim---no approximation, no inspection---and argues that it holds beyond the present substrate. Section 11 draws the consequence for output format, namely that the natural product of a fixed-dimension algebraic memory is a family of representations indexed by resolution rather than a single one. Section 12 extends the substrate to Pearl's ladder, closing the loop opened in Section 1.5. Section 13 states the reservations and the testable predictions. Section 14 concludes by carrying two consequences out of the paper separately from the architecture that produced them. Appendix A distinguishes the kind of equivalence claimed here from mathematical isomorphism, which is not claimed.

\section{Marcus's three pillars and the open questions}

\subsection{The three pillars}

\citet{Marcus2001} reconstructs symbol manipulation as three separable hypotheses, each of which can stand or fall independently. We restate them in the form most relevant to the comparison developed below.

\emph{Pillar 1 (Chapter 3): Operations over variables.} The mind represents abstract relationships between variables and can freely generalize those relationships to novel instances of the variables. The canonical example Marcus develops is the inflection rule that combines a verb stem with the suffix \emph{-ed}: the rule applies to any verb, including verbs not previously encountered, because it operates on the variable \emph{stem} rather than on a list of stored input--output pairs. The technical content of the pillar is that there must exist an operation $f(x)$ that is defined uniformly over all instantiations of the variable $x$, not merely over a stored finite sample. Marcus calls these \emph{universally quantified one-to-one mappings} (UQOTOMs). Multilayer perceptrons trained by backpropagation, in the standard many-nodes-per-variable encoding, do not naturally support free generalization to instantiations outside the trained subspace; this is the central negative result of Chapter 3.

\emph{Pillar 2 (Chapter 4): Structured representations.} The mind has ways of representing complex structures composed of constituent parts, in such a way that the structural relationships among the constituents are preserved and recoverable. Marcus's stock example is the syntactic contrast between \emph{the book that is on the table} and \emph{the table that is on the book}: the constituents are identical, the roles are identical, the structural relationship differs, and the difference is recoverable from the representation. Tensor product representations \citep{Smolensky1990}, circular convolution holographic reduced representations \citep{Plate1995}, temporal synchrony or asynchrony schemes \citep{Shastri1993,Love1999}, and conjunctive coding all offer partial answers; Marcus argues in Chapter 4 that each has a specific weakness, and proposes the treelet as an alternative.

\emph{Pillar 3 (Chapter 5): Individuals distinct from kinds.} The mind distinguishes mental representations of specific individuals (Felix the cat, the chair I am sitting on) from mental representations of the kinds those individuals instantiate (cat, chair). Marcus argues that this distinction is required to support object permanence, to track individuals through time, and to represent propositions involving particular entities rather than category-level generalizations. He notes that many-nodes-per-variable encodings collapse the distinction, because two activations of the same node pattern cannot be distinguished as the same individual encountered twice versus two distinct individuals with similar properties.

\subsection{What the three pillars leave open}

Marcus is explicit about what his program does not settle. Three open questions, each foregrounded in his book, are the targets of the engineering substrate we describe in the following sections. It is worth noting that all three are questions about \emph{substrate}, not about cognition: Marcus had settled to his own satisfaction what the mind does, and was asking what would have to be true of the tissue.

\emph{Open question A: the algebra of operations on registers.} Marcus's treelet proposal (Chapter 4.4) is a proposal for how hierarchical structures can be implemented in a system of registers; it is explicitly not a proposal about the elementary operations the registers support. The treelet supplies the structural scaffold within which constituents are bound to roles, but the algebra under which Bind, Unbind, and recursive composition are defined is left to the implementation. As we develop in Section 3, this open question has a clean answer in GF(2): XOR-and-shift is the unique elementary operation over GF(2) that is reversible, non-commutative under shift composition, and computable as a single bitwise transformation. The treelet is, on this reading, a register set whose elementary operations are XOR-and-shift, and the algebra of the treelet is the algebra of GF(2).

\emph{Open question B: the source of structural specificity without a blueprint.} In Chapter 6.3, Marcus addresses the apparent paradox between evidence that the brain has innately structured microcircuitry and evidence that the DNA does not specify a point-by-point wiring diagram. His proposal is that cascades of master control genes, combined with activity-dependent local signaling, can produce finely specified microcircuitry without a blueprint, much as cascades of regulatory genes produce the heart vasculature \citep[citing][]{Gerhart1997}. He acknowledges that the precise developmental mechanism remains to be identified. The companion paper \citep{Chuma2026b} argues that the DG--CA3 circuit supplies one such mechanism in the form of developmentally specified mossy-fiber target identity \citep{Galimberti2006}, which in the engine framework corresponds to the per-block selection of a primitive generator polynomial $G_b(x)$.

\emph{Open question C: what a register would look like in tissue.} Marcus suggests in Chapter 3.3.4 that registers could in principle be implemented as inter- or intracellular circuits, but he emphasizes that the implementation is open. The companion paper argues that the granule-cell layer of the dentate gyrus, with its theta--gamma clocked sparse binary activation under perforant-path drive, supplies a substrate-level realization of the clocked sparse binary register state that the engine requires. The register, on this reading, is a sparse pattern of granule-cell activity in a clocked behavioral window.

These three open questions structure the development that follows. The engineering substrate of Section 3 answers question A; the biological reading of Section 8 answers questions B and C. Table~\ref{tab:pillars} states the pillar-to-operation correspondence that Sections 4--6 defend one row at a time.

\begin{table}[t]
\centering
\caption{Correspondence between Marcus's three pillars and PyVaCoAl/VaCoAl's operational commitments. The right column gives the elementary operation by which the engine supports the pillar; the operation is reversible in silicon, approximately reversible under sparse-coding biological noise. The associated open question from Marcus's program is given in the rightmost column.}
\label{tab:pillars}
\begin{tabular}{p{4.1cm}p{6.0cm}p{4.4cm}}
\toprule
\textbf{Marcus pillar} & \textbf{Engine operation} & \textbf{Open question addressed} \\
\midrule
1. Operations over variables (Ch. 3) & $\mathrm{Bind}(R,F) = R \oplus \mathrm{shift}(F)$; substitution by re-bind under any filler & Algebra of registers (Q. A) \\
\addlinespace
2. Structured representations (Ch. 4) & $\mathrm{Repr} = \bigoplus_i \mathrm{Bind}(R_i, F_i)$; non-commutative under shift permutation; bit-exact unbind & Algebra of registers (Q. A); composition under fixed dimension \\
\addlinespace
3. Individuals vs. kinds (Ch. 5) & Block-wise address space; distinct hypervectors per individual under reversible kind-binding $\mathrm{Bind}(\text{ISA}, \text{Kind})$ & Storage of new individuals without retraining (related to Q. C) \\
\bottomrule
\end{tabular}
\end{table}

\section{The PyVaCoAl/VaCoAl substrate}

\subsection{Engine commitments at a glance}

PyVaCoAl/VaCoAl \cite{Chuma2026a} is an HDC architecture distinguished from prior HDC and from connectionist embeddings by three operational commitments. We state them here at the level needed for the comparison with Marcus's pillars; full mathematical development is in the companion engineering paper.

\emph{Commitment 1: XOR-and-shift is the unique elementary compute primitive.} Every operation in the architecture---diffusion, binding, unbinding, bundling---decomposes into XOR-and-shift over GF(2). The elementary primitive is the same at every scale of the computation.

\emph{Commitment 2: Diffusion is deterministic and reversible.} For a primitive polynomial $G(x) \in \mathrm{GF}(2)[x]$ of degree $m$, the diffusion operation
\begin{equation}
  \Psi(P) = x^m P(x) \bmod G(x)
  \label{eq:diffusion}
\end{equation}
is implemented as a sequence of $m$ shift-and-conditional-XOR steps through a linear-feedback shift register. $\Psi$ is GF(2)-linear, bijective on the non-collision input space, and satisfies the quasi-orthogonal diffusion (QOD) property: for two distinct inputs whose XOR difference is not in the collision kernel, the output Hamming distance concentrates near $m/2$ with variance $\approx m/4$, exponentially tight in $m$ \citep[Appendix A]{Chuma2026b}.

\emph{Commitment 3: Readout is block-wise majority voting under bounded confidence.} The hypervector of length $L$ is partitioned into $N$ blocks of length $q = L/N$, each block governed by its own primitive polynomial $G_b(x)$ and seed. Block-wise readout produces per-step Confidence Ratio $\mathrm{CR1}$ and path-integral Confidence Ratio $\mathrm{CR2}(n) = \prod_i \mathrm{CR1}(i)$ \citep[Eqs. 4--5]{Chuma2026b}, providing graded readout under partial collision.

The three commitments are not independent design decisions taken in sequence. Commitment 1 fixes the algebra; Commitment 2 is what that algebra does when iterated $m$ times inside a block; Commitment 3 is what a bank of such blocks does when their outputs are counted. There is one decision in this architecture, and the rest is arithmetic.

\subsection{Why XOR-and-shift and not multiplication, convolution, or addition}

The architectural choice of XOR-and-shift as the elementary primitive is not a stylistic preference; it is forced by what Marcus's pillars demand of an elementary operation. We make this explicit because the comparison with prior HDC schemes turns on it, and because the necessity claim of Section 10 is a claim about this paragraph and not about the engine as a whole.

GF(2) admits exactly two non-trivial binary operations: XOR (the field's addition) and AND (the field's multiplication). XOR is its own inverse and is reversible; AND destroys information and is not invertible from outputs alone. Reversibility is what licenses the bit-exact recovery of constituents from a bundled compositional representation---the operational property Marcus most wanted Pillar 2 to support, and the property that defeats circular convolution \citep[whose unbind is a quasi-inverse with accumulating noise; see][]{Plate1995}. There is no third candidate to weigh; in a binary field the choice is between the operation that keeps information and the operation that discards it.

Shift---multiplication by $x$ modulo $G(x)$---supplies what XOR alone cannot: non-commutativity. The bind operation
\begin{equation}
  \mathrm{Bind}(R, F) = R \oplus \mathrm{shift}(F)
  \label{eq:bind}
\end{equation}
distinguishes $\mathrm{Bind}(R, F)$ from $\mathrm{Bind}(F, R)$ because shift acts asymmetrically on the two arguments. This non-commutativity is what allows the bundle of role-filler bindings to distinguish role-asymmetric structures---exactly the property Marcus stressed in the \emph{the dog bites the man} versus \emph{the man bites the dog} contrast (Chapter 4.3). Note what the asymmetry is not: it is not a learned weighting, not a positional embedding, and not a timing difference. It is a permutation, structurally present whether or not anything has been trained, and it costs one shift.

The avalanche property of $\Psi$---a single-bit input perturbation produces an output whose Hamming weight concentrates near $m/2$---is the third architectural consequence of the choice. It is strictly stronger than what random sparse projection of equivalent dimension can offer: random sparse projection with row-sparsity $k$ bounds single-bit-flip output distance at $k$, leaving the bounded-failure mode in which similar inputs produce similar outputs even after projection. The Galois-field diffusion eliminates this failure mode at the level of the elementary operation rather than statistically over many projections. The distinction between eliminating a failure mode structurally and suppressing it statistically will recur; it is the same distinction, in a different register, as the one drawn in Section 1.5 between an operation and a record of outcomes.

\begin{figure}[t]
\centering
\begin{tikzpicture}[
  block/.style={draw, rounded corners, minimum width=4.2cm, minimum height=1.1cm, font=\small, align=center, fill=blue!8},
  primitive/.style={draw, rounded corners, minimum width=2.3cm, minimum height=0.8cm, font=\small, align=center, fill=red!15},
  arrow/.style={-{Stealth[length=2mm]}, thick},
  label/.style={font=\bfseries\small},
  font=\small
]
  \node[label] at (0,2.6) {Three engine commitments};

  \node[block] (c1) at (-5,1.0) {\textbf{C1.} XOR-and-shift\\ as elementary primitive};
  \node[block] (c2) at (0,1.0)  {\textbf{C2.} Primitive-polynomial\\ LFSR diffusion (reversible)};
  \node[block] (c3) at (5,1.0)  {\textbf{C3.} Block-wise\\ majority readout};

  \node[primitive] (xor) at (0,-1.0) {XOR \, $\oplus$};
  \node[primitive] (sh)  at (0,-2.2) {shift \, $\cdot x \bmod G(x)$};

  \draw[arrow] (c1.south) -- (xor.north);
  \draw[arrow] (c2.south) -- (xor.north);
  \draw[arrow] (c3.south) -- (xor.north);

  \draw[arrow] (xor.south) -- (sh.north);

  \node[align=center, font=\scriptsize] at (0,-3.2)
    {\emph{Every operation in the architecture---diffusion, binding,}\\
     \emph{unbinding, bundling---decomposes into these two.}};
\end{tikzpicture}
\caption{The three operational commitments of PyVaCoAl/VaCoAl and their shared elementary primitive. XOR is the unique non-trivial reversible operation on GF(2); the shift gives the operation non-commutativity. All higher-level architectural operations are sequences of these two primitives at different scales.}
\label{fig:engine-commitments}
\end{figure}

\subsection{The elementary operation is the whole argument}

Everything that follows in Sections 4 through 6 is bookkeeping on Equation~\ref{eq:bind}. We flag this now because the three pillars are conventionally treated as three problems, and a reader expecting three solutions will be looking for machinery that is not there.

Pillar 1 asks for an operation defined independently of the value a variable holds; Equation~\ref{eq:bind} is such an operation because $F$ appears only as an argument. Pillar 2 asks that structures built from identical constituents be distinguishable and decomposable; Equation~\ref{eq:bind} distinguishes them because shift is asymmetric, and decomposes them because XOR is an involution. Pillar 3 asks that individuals be separable from kinds without a second representational format; Equation~\ref{eq:bind} separates them because distinct roles are distinct keys into a shared space. One equation, three pillars, no additional apparatus.

The parsimony is worth pausing on, because it is the only evidence of the kind that this paper can offer. We cannot show that GF(2) is the algebra the mind uses; nobody can, on present methods. What we can observe is that a specification written in 2001 without reference to any hardware, listing three requirements its author took to be logically independent, turns out to be satisfiable by one operation chosen for reasons that had nothing to do with cognition---and that the operation is the only one its field admits. That is not proof. It is the kind of coincidence that usually indicates one is looking at a structural fact rather than an engineering accident, and Appendix A states carefully how much weight it will bear.

\section{Pillar 1: Operations over variables as reversible XOR binding}

\subsection{What Marcus asked for}

Marcus's Pillar 1 (Chapter 3) requires an operation that is defined uniformly over all instantiations of a variable. The hallmark of such an operation is \emph{free generalization}: an inflection rule learned on a finite training set extends without retraining to novel verbs, including pseudoverbs the system has never encountered. The standard many-nodes-per-variable multilayer perceptron with backpropagation does not exhibit this property in the general case; this is the central negative result of Marcus's Chapter 3 case studies on linguistic inflection.

What Marcus's positive proposal requires of an operation is, more precisely: (i) the operation must be definable on a variable independently of which value the variable currently holds; (ii) the operation must yield the same structural transformation regardless of the value substituted; (iii) the operation must be inspectable, in the sense that the relationship between variable and value can be recovered from the operation's output. These three desiderata are what we mean by ``variable binding'' in the strong sense Marcus's pillar requires.

\subsection{What VaCoAl's Bind supplies}

The VaCoAl bind operation, Eq.~\ref{eq:bind}, satisfies all three desiderata.

Desideratum (i)---operation independent of value---is satisfied by linearity: $\mathrm{Bind}(R, F_1 \oplus F_2) = \mathrm{Bind}(R, F_1) \oplus \mathrm{Bind}(R, F_2)$, so the binding of the role $R$ commutes with substitution of any value in the filler argument. The operation's definition does not depend on what $F$ currently is.

Desideratum (ii)---structural transformation invariant across substitutions---is the statement that the operation $F \mapsto \mathrm{Bind}(R, F) = R \oplus \mathrm{shift}(F)$ is a fixed function of $F$, parameterized only by the role hypervector $R$. The same role $R$ can be bound to any of an unbounded set of fillers $F$, including fillers not previously encountered, with the same algebraic effect.

Desideratum (iii)---inspectability---is what defeats holographic reduced representations and learned embeddings: it requires that the binding be invertible. Because XOR is its own inverse and the shift operation is bijective (its inverse is the shift-down operation), the unbinding operation
\begin{equation}
  \mathrm{Unbind}(B, R) = \mathrm{shift}^{-1}(B \oplus R) = F
  \label{eq:unbind}
\end{equation}
recovers the filler $F$ exactly under noiseless conditions, and approximately under sparse-coding biological noise. The role $R$ acts as a key; the bind/unbind pair returns the identity at the algebraic level.

This is, in the precise sense Marcus's pillar requires, a substitution-friendly variable binding operation. Given a role $\mathbf{Stem}$ and a filler $\mathbf{walk}$, the bound representation $\mathrm{Bind}(\mathbf{Stem}, \mathbf{walk})$ can be unbound by querying with $\mathbf{Stem}$ to recover $\mathbf{walk}$, and the same operation works for $\mathrm{Bind}(\mathbf{Stem}, \mathbf{blicket})$ where \emph{blicket} is a verb the architecture has never previously encountered. Free generalization is built into the algebra.

\subsection{Why this is stronger than function approximation}

A multilayer perceptron trained on $\{(\mathbf{walk}, \mathbf{walked}), (\mathbf{look}, \mathbf{looked}), \ldots\}$ does not, by virtue of its function-approximator capacity, generalize to $(\mathbf{blicket}, \mathbf{blicketed})$. As \citet{Marcus2001} argues in Chapter 3.2 and developed at length in Chapter 3.5, the issue is not the universal-function-approximator theorem; it is that the function being approximated must be defined on the relevant input subspace, and a perceptron trained on a finite vocabulary has no principled reason to extrapolate to novel inputs in that subspace. The operation over variables, in contrast, is defined on the entire input space by construction: substitution of $\mathbf{blicket}$ for $\mathbf{walk}$ in $\mathrm{Bind}(\mathbf{Stem}, \cdot)$ requires no retraining because the operation is uniform.

The distinction is the one drawn in Section 1.5, and it is worth stating in its sharpest form. A trained approximator holds a record of outcomes; an algebra holds an operation. Records can be extended, and extending them is what scale does. Operations cannot be extended, because there is nothing to extend---they were already defined everywhere. This is why the gap between the two does not close as the corpus grows, and why reporting that a larger model inflects more pseudoverbs correctly answers a question Marcus was not asking.

This is the precise sense in which Pillar 1 distinguishes a symbol-manipulating architecture from a connectionist function approximator. PyVaCoAl/VaCoAl supports the operation as a primitive of the architecture, not as an emergent property of training.

\section{Pillar 2: Structured representations as non-commutative compositional bundling}

\subsection{What Marcus asked for}

Marcus's Pillar 2 (Chapter 4) demands that the mind represent complex structures whose constituent parts and structural relations are jointly recoverable. The book/table contrast he develops in Chapter 4.3---\emph{the book that is on the table} versus \emph{the table that is on the book}---is the canonical diagnostic: the constituents are identical, but the structures must be distinguishable, and the difference must be recoverable from inspection of the representation. The soap-opera example from Chapter 7 develops the same point at higher arity: Amy loves Billy, Billy loves Clara, Clara loves David, $\ldots$, Henry loves Amy---a chain of role-asymmetric bindings whose ironic closure requires representing each binding as a distinct relational fact while keeping all constituents available for further composition.

What Pillar 2 demands of a representational scheme is, more precisely: (i) the same constituents in different structural positions must yield distinguishable representations; (ii) the representation of a compound structure must support recovery of each constituent in each of its roles; (iii) the recursion must continue without exponential growth in representation size, because biological substrates do not have unbounded representational resources.

Marcus's Chapter 4 reviews four proposals that fail one or more of these desiderata. Conjunctive coding satisfies (i)--(ii) but explodes combinatorially in (iii). Tensor products \citep{Smolensky1990} satisfy (i)--(ii) but require representations whose dimensionality grows multiplicatively with the depth of embedding, with five-level structures requiring on the order of $10^7$ nodes under reasonable per-role precision \citep[Chapter 4.3.3]{Marcus2001}. Circular convolution holographic reduced representations \citep{Plate1995} keep the dimensionality fixed but at the cost of replacing exact unbind with a noisy quasi-inverse, accumulating crosstalk that limits recursion depth in practice. Temporal synchrony and asynchrony schemes \citep{Shastri1993,Love1999} satisfy (i)--(iii) under idealized timing assumptions but face a combinatorial-precision problem that Marcus develops in detail (Chapter 4.3.4).

\subsection{What VaCoAl's bundling supplies}

A bundle of role-filler bindings in PyVaCoAl/VaCoAl is the XOR sum
\begin{equation}
  \mathrm{Repr} = \bigoplus_{i=1}^k \mathrm{Bind}(R_i, F_i) = \bigoplus_{i=1}^k \left( R_i \oplus \mathrm{shift}(F_i) \right).
  \label{eq:bundle}
\end{equation}
We claim that this satisfies desiderata (i)--(iii) under the algebraic properties of Section 3.

Desideratum (i)---same constituents in different structural positions yield distinguishable representations---is satisfied by the non-commutativity of bind. For two bindings $\mathrm{Bind}(R_1, F_1) \oplus \mathrm{Bind}(R_2, F_2)$ and $\mathrm{Bind}(R_1, F_2) \oplus \mathrm{Bind}(R_2, F_1)$, the constituents $\{F_1, F_2\}$ and roles $\{R_1, R_2\}$ are identical but the bundles differ because shift acts asymmetrically on the arguments. The book/table contrast is recovered: $\mathrm{Bind}(\mathbf{Inner}, \mathbf{Book}) \oplus \mathrm{Bind}(\mathbf{Outer}, \mathbf{Table})$ differs from $\mathrm{Bind}(\mathbf{Inner}, \mathbf{Table}) \oplus \mathrm{Bind}(\mathbf{Outer}, \mathbf{Book})$ at the level of the bundled hypervector, despite identical multiset structure on roles and fillers.

Desideratum (ii)---per-constituent recovery in role---is satisfied by the algebraic Unbind operation. For the bundle in Eq.~\ref{eq:bundle}, querying with role $R_j$ returns
\begin{equation}
  \mathrm{Unbind}(\mathrm{Repr}, R_j) = \mathrm{shift}^{-1}\!\left( \mathrm{Repr} \oplus R_j \right) = F_j \oplus \xi_j,
  \label{eq:unbind-bundle}
\end{equation}
where $\xi_j$ is the bundled crosstalk from the other $k-1$ bindings. Under the QOD property of $\Psi$ (Section 3.1), $\xi_j$ has expected Hamming weight near $m/2$ and is effectively orthogonal to $F_j$ in the cleanup stage. Block-wise majority voting against a candidate vocabulary then recovers $F_j$ with confidence $\mathrm{CR1}$ that approaches unity in the no-collision regime.

The contrast with circular convolution is worth stating exactly, because the two situations look similar from a distance and are not. In both cases the query returns the target plus something. The difference is where the something lives. In circular convolution the residual is bona fide noise in the same algebra as the signal: it is additive in the representational space, it accumulates with bundle size, and no downstream stage can distinguish it from a contribution to the answer. In Eq.~\ref{eq:unbind-bundle} the residual is scattered across the address space by the avalanche property, which is to say it does not accumulate anywhere in particular, and the contractive readout removes it. A structurally scattered residual and an algebraically accumulated one are not two amounts of the same quantity. They are different objects, and Section 10 returns to the point.

Desideratum (iii)---fixed-dimensional recursion---is satisfied by linearity. The bundle of bindings is itself a hypervector in the same space as its constituents, and can be bound again as a filler in a higher-order role-filler binding. The five-level embedding that required $10^7$ nodes in tensor product calculus uses the same fixed $L$-bit hypervector in VaCoAl, with capacity limited by collision saturation rather than by dimension growth. Concretely, with $L = 1000$ and $N = 100$ blocks, the architecture supports recursive composition to depth approximately $\log_2(N) \approx 7$ before block collisions saturate, exceeding the depth at which human syntactic embedding becomes psycholinguistically unmanageable.

That last coincidence deserves neither more nor less weight than it can bear. We do not claim that the architecture predicts the psycholinguistic limit; the two numbers are close, the mechanisms are not shown to be the same, and $N$ is a free parameter. What can be said is that the engine's failure mode is of the right \emph{kind}: it degrades by running out of distinguishable addresses at depth, which is a graded and reportable failure, rather than by requiring $10^7$ nodes, which is not a failure mode at all but an impossibility.

\subsection{The soap opera worked out}

The Amy-loves-Billy chain from \emph{The Algebraic Mind}, Chapter 7, is a useful test. Each binding is
\begin{equation}
  L_i = \mathrm{Bind}(\mathbf{Lover}, \mathbf{Person}_i) \oplus \mathrm{Bind}(\mathbf{Beloved}, \mathbf{Person}_{i+1}),
\end{equation}
yielding an 8-fact bundle
\begin{equation}
  S = \bigoplus_{i=1}^8 L_i, \qquad \mathbf{Person}_9 = \mathbf{Person}_1 = \mathbf{Amy}.
\end{equation}
The ironic closure---that Henry loves Amy, completing the cycle---is recovered by querying $\mathrm{Unbind}(S \oplus L_{1\ldots 7}, \mathbf{Lover}) = \mathbf{Henry}$ and $\mathrm{Unbind}(S \oplus L_{1\ldots 7}, \mathbf{Beloved}) = \mathbf{Amy}$, with $\mathrm{CR2}(2) = \mathrm{CR1}(\text{Henry}) \cdot \mathrm{CR1}(\text{Amy})$. The architecture supports the rapid binding that Marcus emphasizes the soap-opera example requires, without retraining and at fixed dimensionality.

Two features of this worked example are easy to pass over. The first is that the closure is recovered by \emph{subtraction}---$S \oplus L_{1\ldots 7}$ removes seven facts from the bundle exactly, leaving the eighth. Subtraction of a constituent is available only where composition was exact; in a lossy scheme one can add facts to a bundle but never cleanly remove one, and the eighth fact would have to be inferred from what remains rather than read off it. The second is that the confidence attached to the answer, $\mathrm{CR2}(2)$, is a product of per-step margins rather than a single similarity score. It is a statement about the route by which the answer was reached, not about the answer alone, and it is the same quantity that Section 12 identifies as the engine's contribution to Pearl's third rung.

\section{Pillar 3: Individuals versus kinds under reversible indexing}

\subsection{What Marcus asked for}

Marcus's Pillar 3 (Chapter 5) requires that the mind distinguish representations of specific individuals from representations of the kinds those individuals instantiate. The desideratum is not merely lexical---a node labelled \texttt{Felix} versus a node labelled \texttt{cat}---but operational: the architecture must support the assertion that \emph{Felix is a cat} as a fact distinct from the assertion that \emph{Felix has fur, four legs, and a tail}, and the assertion that \emph{Felix} encountered today is the same Felix encountered yesterday, despite possible differences in observed properties. Marcus argues in Chapter 5.1 that many-nodes-per-variable encodings collapse this distinction because two activations of the same node pattern cannot, on their own, signal sameness of individual.

What Pillar 3 demands of an architecture is, more precisely: (i) a representational format that supports a fixed, persistent identifier per individual, separate from the individual's current observed properties; (ii) a binding mechanism that ties the individual to its kind without collapsing the two into a single undifferentiated representation; (iii) a substrate for adding new individuals to memory without retraining or rewiring the representations of existing individuals.

\subsection{What VaCoAl supplies}

PyVaCoAl/VaCoAl supports each of these through the same algebra developed in Sections 3--5, with the addition of an explicit index-vector convention.

Desideratum (i)---persistent individual identifiers---is supplied by Entry Addresses (EAs). Each individual receives a distinct hypervector identifier at first encoding, drawn from a high-dimensional space in which the QOD property guarantees that two random EAs are quasi-orthogonal with overwhelming probability. The EA persists across encounters with the individual; what varies across encounters is the set of bindings that link the EA to currently observed properties.

Desideratum (ii)---kind binding distinct from property bundling---is supplied by an ISA role:
\begin{equation}
  \mathrm{Kinds}(\mathrm{EA}_{\mathrm{Felix}}) = \mathrm{Bind}(\mathbf{ISA}, \mathbf{Cat}),
\end{equation}
while observed properties are a separate bundle
\begin{equation}
  \mathrm{Props}(\mathrm{EA}_{\mathrm{Felix}}) = \mathrm{Bind}(\mathbf{HasColor}, \mathbf{Orange}) \oplus \mathrm{Bind}(\mathbf{HasAge}, \mathbf{Five}) \oplus \cdots
\end{equation}
The kind binding and the property bundle live in the same representational space and use the same algebra, but they are addressable separately because the queries $\mathbf{ISA}$ and $\mathbf{HasColor}$ are distinct role hypervectors. The architecture distinguishes \emph{Felix is a cat} (a kind binding) from \emph{Felix is orange and aged five} (a property bundle) at the operational level---not by virtue of separate representational substrates, but by virtue of the queries used to recover each.

Desideratum (iii)---adding new individuals---is satisfied by the structurally trivial operation of drawing a fresh EA from the quasi-orthogonal address space and writing it. No existing EA is disturbed; no representation is retrained; the architecture's capacity for new individuals is bounded only by collision saturation in the address space. This is the engineering counterpart of the biological observation, developed in the companion paper, that adult neurogenesis in the dentate gyrus supplies a continuing supply of fresh granule cells with new mossy-fiber connectivity \citep{Kempermann2018,Akers2014}---a substrate-level realization of fresh-EA generation.

It is worth registering how much of Marcus's Chapter 5 worry this dissolves and how much it does not. What it dissolves is the collapse: two encounters with the same EA are the same individual by construction, and two distinct EAs are distinct individuals however similar their property bundles, because the identifier is not derived from the properties. What it does not dissolve is the tracking problem---deciding, on encountering an orange cat, whether to retrieve $\mathrm{EA}_{\mathrm{Felix}}$ or to mint a fresh one. That decision is a retrieval question, answered by the readout with confidence $\mathrm{CR1}$, and it is fallible in exactly the way one would expect. The architecture guarantees that individuals \emph{can} be kept distinct; it does not guarantee that they always \emph{will} be, and no architecture that must sometimes recognise things could.

\subsection{The kind/individual asymmetry under the same algebra}

A subtle but important point: kind hypervectors and individual hypervectors are drawn from the same space and obey the same algebra. The asymmetry between them is not representational but \emph{relational}: an individual's EA is linked to a kind through an ISA binding, but a kind is linked to its members through an ensemble of inverse bindings (or, more economically, through the kind's role as filler in many individuals' kind bindings). The distinction Marcus's Pillar 3 demands is thereby supported by the architecture without requiring two distinct representational formats. The same algebra that supports Pillars 1 and 2 supports Pillar 3 through the use of indexing rather than through the introduction of new machinery---a parsimony that, we suggest, was not available to the 2001 alternatives.

\section{Treelets reinterpreted: algebraic register sets indexed by primitive polynomials}

\subsection{Marcus's treelet, revisited}

Marcus's Chapter 4.4 introduces the treelet as a preorganized, hierarchical arrangement of register sets. Each register set holds the encoding for a simple element---\textbf{cat}, \textbf{dog}, \textbf{Mary}, \textbf{love}, \textbf{blicket}---in a fixed number of registers, with the encoding stable across reuse. The mind is hypothesized to have on hand a large stock of empty treelets, with new knowledge represented by filling in an empty treelet with values.

Marcus is explicit about what the proposal does and does not commit to. It is a proposal about how hierarchical structures can be implemented in a system of registers; it is not a proposal about how registers themselves are implemented in neurons, nor a proposal about the elementary operations the registers support. These two open questions---the algebra of register operations, and the neural realization of the register---are exactly the two questions we have been pursuing.

\subsection{The block as treelet, the polynomial as treelet identity}

The reinterpretation we propose is that a treelet is an algebraic register set whose elementary operations are XOR-and-shift over GF(2), and whose identity---what distinguishes one treelet from another in the stock---is the primitive generator polynomial $G_b(x)$ that governs its diffusion operation. Concretely:

\emph{Register $\to$ block.} In Marcus's terms, a register holds the encoding of a simple element. In VaCoAl, a block of the hypervector, indexed $b \in \{1, \ldots, N\}$, holds an $m$-bit sub-vector under primitive polynomial $G_b(x)$. The block is the register; the bits of the block are the individual registers in the set.

\emph{Treelet $\to$ tuple of blocks under shared polynomial structure.} In Marcus's terms, a treelet is a hierarchical arrangement of register sets that jointly encode a structured element---a bound role-filler pair, or a recursive composition. In VaCoAl, a treelet corresponds to a tuple of blocks whose diffusion polynomials are jointly chosen to support a common operation: for instance, a tuple $(G_{\mathrm{role}}, G_{\mathrm{filler}})$ supporting a single role-filler binding, or a tuple of $k$ such pairs supporting a $k$-arity relational structure.

\emph{Empty treelet $\to$ block under fresh seed.} In Marcus's terms, the mind has a stock of empty treelets that can be filled in. In VaCoAl, the equivalent is a block whose polynomial $G_b(x)$ is fixed but whose seed register is initialized to zero, awaiting the deposit of an EA at the addressed location.

\emph{Treelet identity stable across reuse $\to$ polynomial $G_b(x)$ fixed across operations.} In Marcus's terms, a treelet's structural identity is independent of the values currently stored in it. In VaCoAl, the polynomial $G_b(x)$ is fixed across all operations on block $b$, even as the values in the block change; the polynomial \emph{is} the treelet's identity.

\begin{table}[t]
\centering
\caption{Reinterpretation of Marcus's treelet (Chapter 4.4) as a PyVaCoAl/VaCoAl algebraic register set. Each row maps a feature of the 2001 treelet proposal to a feature of the engine's block-wise architecture. The right column indicates whether the feature is bit-exact in silicon or approximate under biological noise.}
\label{tab:treelet}
\begin{tabular}{p{5.0cm}p{6.5cm}p{2.5cm}}
\toprule
\textbf{Marcus 2001 treelet} & \textbf{VaCoAl block-wise engine} & \textbf{Realization} \\
\midrule
Register set & Block $b$ of hypervector, $m$ bits & exact \\
Treelet identity & Primitive polynomial $G_b(x)$ & exact \\
Stock of empty treelets & Library of distinct primitive polynomials per block & exact \\
Filling in a treelet & Depositing EA at the addressed block location & exact \\
Treelet identity stable across reuse & $G_b(x)$ fixed across operations on block $b$ & exact \\
Hierarchical composition of treelets & Recursive bundling at fixed hypervector dimension & approximate (collision-saturated) \\
\bottomrule
\end{tabular}
\end{table}

\subsection{What this buys}

The reinterpretation is not merely a renaming. It supplies the 2001 treelet program with two operational commitments it lacked: an explicit algebra for the operations on registers (XOR-and-shift over GF(2)) and an explicit specification of how distinct treelets are distinguished (the choice of $G_b(x)$). The algebra is what enables free generalization (Pillar 1) and bit-exact compositional inspection (Pillar 2). The polynomial-indexed treelet identity is what supplies, in the biological reading, the structural specification that distinguishes one developmentally programmed mossy-fiber target pattern from another (the answer to Marcus's Open Question B).

There is a further consequence worth stating, because it bears on the sense in which the treelet was ever a separate thing. Marcus's stock of empty treelets is a stock of \emph{structures awaiting content}, and the puzzle it raises is where the structures come from if not from the content. On the present reading the puzzle does not arise: the structure of a block is its polynomial, the polynomial is fixed before anything is written, and writing does not modify it. The stock of empty treelets is a library of polynomials---a finite, enumerable, developmentally specifiable object---and the mystery of pre-organized structure reduces to the question of how a developmental program picks generators. That is still an open question, but it is an open question of a much more tractable kind than the one it replaces.

The treelet, on this reading, is not a separate cognitive module; it is the architectural unit of the algebra Marcus's program required.

\section{Neural implementation: DG--CA3 as the biological engine}

\subsection{The five-homologue reading in brief}

The companion Perspective \citep{Chuma2026b} argues that the DG--CA3 circuit instantiates the same engine, with five biophysical homologues corresponding to the five operational commitments of the silicon LFSR. We summarize the reading here at the level needed for the comparison with Marcus's Open Questions B and C; the full argument, including cellular and synaptic evidence, is in the companion paper.

\emph{Component 1: GF(2) register state.} Granule cells in the dentate gyrus fire sparsely (typically below 1--2\% activation; \citealp{Leutgeb2007,Diamantaki2016}) and lock to theta--gamma oscillatory cycles \citep{Pernia2014}. The combination---sparse activation under clocked oscillatory structure---supplies the substrate-level realization of the clocked sparse binary register state that the engine requires. Marcus's Open Question C, ``what would a register look like in tissue,'' is answered by reading the granule-cell ensemble in a clocked behavioral window as a sparse binary register.

\emph{Component 2: XOR compute primitive.} \citet{Gidon2020} demonstrated in human layer 2/3 cortical pyramidal neurons that dendritic calcium action potentials implement XOR-like input--output transformations: maximal response to moderate inputs and suppression of both weak and strong inputs, the canonical XOR signature. \citet{Benavides2020} showed that human hippocampal CA1 pyramidal neurons share the dendritic morphology supporting such nonlinear integration. Sublinear integration in granule-cell dendrites \citep{Krueppel2011} supplies the intermediate regime. The XOR primitive of the engine has a direct biological correspondent in the pyramidal-cell family that constitutes the hippocampal input and output stages.

\emph{Component 3: Deterministic operation propagation.} Mossy-fiber detonator transmission \citep{Henze2002,Vyleta2016,Chamberland2018} ensures that single granule-cell activations reliably propagate to CA3 pyramidal targets, eliminating the bounded-failure mode of random sparse projection at the substrate level rather than statistically.

\emph{Component 4: Structural specification of operations.} The mossy-fiber projection from granule cells to CA3 is sparse, directed, anatomically specified, and developmentally programmed \citep{Acsady1998,Galimberti2006,Rollenhagen2010,Wilke2013}. Each granule cell defines a specific operation over a specific small set of CA3 targets. This is the biological correlate of selecting a generator polynomial $G_b(x)$, and the answer to Marcus's Open Question B: the developmentally programmed mossy-fiber targeting is exactly the kind of innate, non-blueprint microcircuitry that cascading master control genes can produce.

\emph{Component 5: Contractive readout.} CA3 recurrent connectivity at $\sim$1\% supports attractor-based pattern completion under threshold-linear readout \citep{Guzman2016,Rolls2023}. The attractor dynamics collect block-wise outputs into a coherent reading, the biological analogue of block-wise majority voting in silicon.

\subsection{Closing Marcus's open questions}

The biological reading addresses Marcus's three open questions directly. Open Question A (algebra of registers) is answered by the engineering substrate: XOR-and-shift over GF(2). Open Question B (structural specificity without a blueprint) is answered by developmentally specified mossy-fiber target identity, which corresponds in the engine framework to per-block selection of a primitive generator polynomial. The cascading-gene mechanism Marcus proposed in Chapter 6.3 has, in the intervening years, been substantially elaborated for developmental specification of synaptic targeting \citep{Galimberti2006,SudhofRevised2008}, and the resulting target identity is at single-cell resolution. Open Question C (what a register would look like in tissue) is answered by the clocked sparse binary granule-cell ensemble.

We do not claim that the answers are final, and we are careful about the kind of claim being made. This is not an argument that the engine mimics the hippocampus, nor that the hippocampus implements an LFSR. It is an argument of convergent design: two substrates, one carbon and one silicon, faced the same computational problem---orthogonalise sparse inputs, propagate them without loss of identity, and contract many local verdicts into one---and arrived at structurally similar solutions because the problem admits few. Appendix A states this as \emph{structural flow equivalence} rather than mathematical isomorphism, and marks precisely where the weaker claim is all that the evidence supports. Bit-exact reversibility of GF(2) operations holds in silicon but only approximately under biological noise; the correspondence between developmental programs and primitive-polynomial selection is structural rather than proven by functional-equivalence theorems. The honest claim is that the open questions Marcus identified in 2001 now have candidate answers at the level of substrate, and the answers are testable.

\section{What this is, that 2001 alternatives were not}

\subsection{Tensor products}

\citet{Smolensky1990} proposed tensor product representations as the algebraic substrate of structured binding. The operation is the outer product of role and filler vectors: $\mathrm{Bind}(R, F) = R \otimes F$. This satisfies non-commutativity (in general $R \otimes F \neq F \otimes R$) and exact unbind (by contraction). But the dimension of the bound representation is the product of the dimensions of the constituents, and the dimension of a $k$-level recursively bound structure is the product of $k$ such dimensions. Marcus (Chapter 4.3.3) develops the concrete cost: a five-level embedding under reasonable per-role precision requires on the order of $10^7$ nodes.

VaCoAl's bundling occupies the same algebraic role---non-commutative compositional binding with exact unbind---but at fixed hypervector dimension. The depth of recursion is bounded not by dimensional explosion but by collision saturation in the block-wise address space, with capacity scaling logarithmically in the number of blocks. This is the architectural advantage that makes the program operational where the tensor product remained, in Marcus's words, ``not impossible but not terribly plausible either'' (Chapter 4.3.3). Note that the tensor product is not being accused of getting the algebra wrong. It gets the algebra right and cannot afford it; VaCoAl keeps the same two properties and changes the bill.

\subsection{Holographic reduced representations}

\citet{Plate1995} proposed circular convolution as the binding operation, with circular correlation as a quasi-inverse for unbind. The advantage over tensor products is that the dimension of the bound representation equals the dimension of the constituents---fixed-dimensional recursion is achieved. The disadvantage is that the quasi-inverse is not exact: unbind recovers the filler plus a noise term whose variance accumulates with the number of bindings in the bundle, limiting recursion depth in practice and ruling out the bit-exact compositional inspection that Marcus's Pillar 2 most wanted.

PyVaCoAl/VaCoAl preserves the fixed-dimensional recursion of HRR but replaces the quasi-inverse with bit-exact reversibility through XOR-and-shift. The unbind operation is the algebraic inverse of bind, not its approximation. Crosstalk, when present (in the multi-binding case), is structurally scattered across the address space rather than algebraically accumulated in the signal subspace, and is removed by the contractive readout. The bit-exactness is what licenses, in turn, the reversible-binding reading of Pearl's intervention operator developed in Section 12 below. HRR is the closest of the alternatives and the one whose failure is most instructive, which is why Section 10 is written about it rather than about the others.

\subsection{Temporal synchrony and asynchrony}

\citet{Shastri1993} proposed temporal synchrony---bindings represented by synchronized firing of the bound nodes---as the binding mechanism. \citet{Love1999} proposed temporal asynchrony---bindings represented by asymmetric firing sequences---as an alternative. Marcus (Chapter 4.3.4) develops the worry that both schemes face a combinatorial precision problem: as the number of represented propositions grows, the system requires increasingly fine temporal resolution to distinguish bindings, with thresholds that may exceed biological precision.

PyVaCoAl/VaCoAl replaces the temporal-precision substrate with an algebraic substrate, and the binding asymmetry comes from the shift permutation rather than from firing-time differences. The precision problem is structurally absent: discrimination between bindings is supported by the QOD property of the diffusion operation, which is exponentially tight in $m$ rather than linearly tight in temporal precision. This is a change in how discrimination scales, not an improvement in how carefully it is done---the sort of difference that cannot be recovered by better engineering on the other side.

\subsection{Multilayer perceptron embeddings}

The dominant contemporary approach to relational representation is learned dense embeddings, trained by gradient descent on relational prediction tasks. Marcus's Chapter 3 critique applies: the embeddings do not freely generalize to constituents outside the training subspace, the structural relations are not algebraically inspectable, and the kind/individual distinction is collapsed unless explicitly engineered as a separate representational format. The success of embeddings on large-scale prediction tasks should not be confused with their adequacy as accounts of the Algebraic Mind program; they are pattern associators, in the precise sense of Marcus's Chapter 7, and the program he developed was about what pattern associators cannot do.

We restate the Position claim of Section 1.4 here, because this is where it is most likely to be misread. Nothing in this comparison is an argument that such systems should be replaced, or that they do their own work badly. They do something this substrate does not do at all, and the observation that they do not do symbol manipulation is an observation about division of labour rather than about quality.

\section{No Approximation, No Inspection}

It is worth isolating the general form of what Sections 4--6 show, because it is easy to read the result as a list of three conveniences when it is in fact a single constraint---and one that runs in a direction most architectures would prefer it did not.

Set the elementary operation to a quasi-inverse. Nothing catastrophic happens at first: unbinding returns the filler plus a small residual, retrieval still works, and for shallow structures the difference is imperceptible. But consider what has changed about the \emph{type} of the returned object. Under an exact inverse, $\mathrm{Unbind}(\mathrm{Repr}, R_j)$ is a constituent: it was placed there, it is recovered, and the operation of removing it from the bundle---$\mathrm{Repr} \oplus \mathrm{Bind}(R_j, F_j)$---is defined and exact. Under a quasi-inverse, what is returned is a point near where a constituent would be, and removal is not defined, because subtracting an approximation leaves a residue that is indistinguishable from the remaining content. The first object supports decomposition. The second supports similarity judgment. They are not two grades of the same thing.

Once that is seen, the three pillars stop looking independent. Pillar 2's inspectability is gone directly, by the argument just given. Pillar 1's free generalization goes with it, because substitution requires unbinding at a site, replacing, and re-binding, and each cycle through an approximate inverse compounds---so the operation that was supposed to be uniform over fillers becomes, in practice, an operation whose reliability depends on how many times it has been applied, which is to say a record of outcomes again. Pillar 3 goes last and least visibly: individual and kind remain nominally separable, but the separation is now a threshold on similarity, and two individuals whose property bundles overlap sufficiently will merge under exactly the conditions Marcus's Chapter 5.1 identified. Weaken the elementary operation and all three fail together, in that order, for one reason.

The general statement is this: \emph{compositionality is purchased with decomposability}. A representation that cannot be exactly taken apart was never composed; it was mixed. The ability to build a structure and the ability to inspect one are not two capabilities that an architecture might have in different degrees---they are the same algebraic property described from opposite ends, and an architecture that has one has the other. This is why the deficiency of lossy binding is not a defect of a particular implementation to be designed away in a later version, and why it cannot be repaired downstream: no amount of cleverness at the readout can recover a decomposition that the elementary operation did not preserve.

Two qualifications keep this from being triumphalism. First, the lossy schemes purchase something real with what they give up: graded similarity, available directly at the representational level, which is what makes them natural for perception and for generalisation over noisy inputs. The engine gives this up at the elementary operation and re-acquires it one level higher, as $\mathrm{CR1}$ and $\mathrm{CR2}$ at readout---exactness underneath, gradedness on top. That is a different arrangement of the same two goods, not a free acquisition of both, and there are problems for which the other arrangement is better. Second, the constraint we state is about the algebra, not about tissue. Biological substrates are noisy, and the reversibility we attribute to DG--CA3 in Section 8 holds statistically and under sparse coding. What the constraint says about biology is narrower and, we think, still interesting: to the extent that a neural system does support compositional inspection, something in it must be doing approximately exact decomposition, and it is worth asking what.

\section{Resolution as Part of the Answer}

The capacity analysis of Section 5.2 has an implication for output format that we state separately, because it does not follow from any single configuration and it is easy to file as a limitation.

The engine has two free parameters, the block count $N$ and the memory depth per block $m$, and under a fixed total capacity they trade against each other. Raising $N$ buys orthogonal resolution---more independent local verdicts, finer discrimination among structures---at the cost of shallower per-block address space and therefore more collisions. Raising $m$ buys collision headroom at the cost of resolution. Neither setting is correct. Each yields a representation that is internally coherent, decomposable to a characteristic depth, and honest about its own limits through $\mathrm{CR1}$ and $\mathrm{CR2}$; and the structures that survive to the top of a deep composition are not the same across settings.

The natural output of such an architecture is therefore not a single representation but a family indexed by resolution---closer to a spectrum than to an answer. Reporting that a structure is recoverable to depth $d_1$ at one resolution and to depth $d_2$ at another, together with the parameter that selects between them, is strictly more informative than reporting either alone, and it is a report that a system whose output type is one plausible continuation cannot make. The resolution parameter is not noise to be tuned away before publication; it is part of what was found.

This bears on Marcus's program in a specific way. The psycholinguistic embedding limit---the depth at which centre-embedded structures stop being processable---is usually treated as a fact about human working memory to be explained. On the present reading it is what a resolution parameter looks like from the inside. We do not claim to have explained it; the numbers coincide loosely and the mechanisms are not shown to be the same (Section 5.2). What we claim is narrower: an architecture in which composition depth is a parameter rather than a constant is the kind of architecture in which such a limit could be a design point rather than a defect, and that is a more comfortable place for the fact to sit than the alternatives afford.

\section{Outlook: from the Algebraic Mind to counterfactual reasoning}

\subsection{Why Pearl's ladder is relevant}

Section 1.5 argued that Marcus's pseudoverb test is already a counterfactual query. This section closes that loop.

The Algebraic Mind program as Marcus developed it targets symbol manipulation at the level of operations over variables, structured representations, and individuals. \citet{Pearl2018,Pearl2009} provides a complementary classification of cognitive capacity at the level of causal queries: rung 1 (association) is captured by $P(Y \mid X)$ and answerable from observation; rung 2 (intervention) requires surgical modification of one variable holding others fixed, captured by the do-operator $P(Y \mid \mathrm{do}(X))$; rung 3 (counterfactuals) requires holding the factual world and a contrary-to-fact world simultaneously, captured by $P(Y_x \mid X', Y')$. Pearl's argument that rung 2 is provably unanswerable from observational data alone, and that rung 3 requires parallel non-interfering representations of factual and counterfactual worlds, locates a capacity that no purely pattern-associative architecture supports.

\subsection{Engine commitments and the ladder}

PyVaCoAl/VaCoAl's commitments map onto the ladder as follows; the development is condensed from Appendix B of the companion Perspective \citep{Chuma2026b}.

\emph{Rung 1.} Associative retrieval through block-wise voting with confidence CR1 supplies the rung-1 capability. Both Marcus's program and the engine reading support this; standard HDC and dense embeddings also support it.

\emph{Rung 2.} The do-operator requires surgical modification of a single role binding without disturbing others. The engine's reversible bind/unbind supports this operationally: starting from $\mathrm{Repr}$, the unbind of role $R_j$ to recover $F_j$, the substitution of $F_j$ by a counterfactual $F_j'$, and the rebind to recover the modified $\mathrm{Repr}'$ is a sequence of XOR-and-shift operations that affects only the $R_j$ binding. This is the architectural feature that distinguishes the engine reading from approximate compositional schemes; circular convolution's lossy unbind cannot support surgical intervention without accumulating noise from all other bindings. The word \emph{surgical} is doing real work here, and Section 10 explains why it cannot be approximated: an intervention that perturbs the bindings it was supposed to hold fixed is not a less precise intervention but a different operation.

\emph{Rung 3.} Pearl's counterfactual evaluation requires simultaneous representation of factual and counterfactual worlds. The companion paper argues that the conserved two-orthogonalizer hippocampal architecture---Regime A scaffold for factual representation, Regime B mossy-fiber writes for counterfactual generation---supplies the parallel non-interfering substrate that rung 3 demands. CR2 path traces over the two regimes provide an algorithmic estimator for $P(Y_x \mid X', Y')$.

This is an extension of Marcus's program in one sense and a restatement of it in another. It is an extension because Marcus did not develop the counterfactual capacity in 2001; his program was about symbol manipulation at the level of operations and structures, not at the level of causal queries. It is a restatement because, as Section 1.5 argued, the demand he did make---that a rule extend to an instance never encountered---is already a demand for evaluation along a road not taken, and cannot be met by any system whose representations record only where it has been. The Algebraic Mind and the causal ladder turn out to be two vocabularies for one requirement: that the architecture hold operations rather than outcomes. What the engine adds is a substrate on which the requirement is met by construction rather than hoped for.

\section{Reservations and testable predictions}

\subsection{Reservations}

Three reservations bound the strength of the correspondence developed above.

First, bit-exact reversibility of GF(2) operations holds in silicon but only approximately under biological noise. The companion paper develops this reservation in detail (Section 9.1). Long-term potentiation and long-term depression are stochastic, recall under interference is graded, and population-level reversibility we attribute to the DG--CA3 circuit holds only in a statistical sense that requires sparse coding to be operative. Plate's HRR faces the same limitation relative to its idealized algebraic form; the engine reading inherits it. The reservation does not undermine the correspondence with Marcus's pillars at the architectural level---the algebra is the algebra---but it bounds the strength of any claim about exact biological realization. It also interacts with Section 10 in a way we should not paper over: if biological binding is only approximately reversible, then the necessity argument implies that biological compositional inspection is correspondingly bounded, which is an empirical prediction and not obviously a comfortable one.

Second, the biological correlate of the choice of primitive generator $G_b(x)$ remains unidentified at the molecular level. Mossy-fiber target specificity is developmentally programmed, but whether the developmental program is functionally equivalent to selecting a primitive generator over GF(2) is an open empirical question. The correspondence is structural; the functional equivalence is conjectural.

Third, the treelet reinterpretation we develop in Section 7 is one reading of Marcus's proposal, not the only possible reading. Marcus's treelet is general enough that other algebras could in principle be assigned to it. We argue that GF(2) is the most parsimonious choice---it satisfies the reversibility, non-commutativity, and free-generalization desiderata under a single elementary primitive---but the argument is one of parsimony, not necessity. Section 3.3 states what we take parsimony to be worth here, and it is less than proof.

\subsection{Testable predictions}

The correspondence generates falsifiable predictions at three levels.

\emph{Architectural-level prediction (engineering).} If PyVaCoAl/VaCoAl is the correct algebra for Marcus's pillars, then concrete cognitive tasks that are diagnostic for Marcus's program---linguistic inflection generalization to pseudoverbs, soap-opera role-filler tracking, kind/individual distinction in object permanence---should be solvable on the architecture with capacity scaling logarithmically in block count and confidence given by CR1/CR2. Failure modes should be predictable from collision saturation rather than from training-data coverage. Direct implementation of Marcus's Chapter 3 and Chapter 5 case studies in PyVaCoAl is a concrete test, and we state plainly that it has not yet been run.

\emph{Biological-level prediction (companion paper).} The four predictions developed in the companion Perspective \cite{Chuma2026b}--[Section 8]---multiplicative decay in iEEG multi-hop replay, absence of random-projection bounded-failure mode at minimum-input perturbations, primate-versus-rodent per-cell pattern-separation efficiency, and quantitative encoding/retrieval lesion dissociation---constrain the biological reading independently of the engineering substrate. Each is testable with current methods.

\emph{Cognitive-level prediction (Marcus's program).} If the substrate supports counterfactual reasoning as Section 12 argues, then lesions selectively impairing Regime B should impair counterfactual reasoning tasks more severely than rung-1 associative retrieval, and the dissociation should follow the multiplicative CR2 decay curve of the engine reading. Existing cognitive-neuropsychology paradigms on counterfactual reasoning in patients with selective hippocampal damage \citep{Mullally2014} are a candidate test bed, though the methodology would need to be refined to dissociate Regime A and Regime B contributions.

\emph{A prediction that would refute the necessity claim.} Section 10 asserts that the three pillars stand or fall with the exactness of the elementary operation. The claim is refutable in the most direct way available: an architecture that supports bit-exact compositional inspection on top of a demonstrably lossy elementary binding operation would falsify it. We do not think one exists, but the burden is ours, and this is where it should be discharged.

\section{Conclusion}

\emph{The Algebraic Mind} closed with the statement that, even granting the three pillars of symbol manipulation, the work of discovering how those pillars are implemented in neural hardware remained to be done. The book's positive proposals---registers as substrates for variable values, treelets as hierarchical arrangements of registers, cascading master control genes as the developmental mechanism for innately structured microcircuitry---were explicitly acknowledged as plausible hypotheses rather than proven facts.

Twenty-five years later, the substrate Marcus's program called for is available, in the form of an algebro-deterministic HDC architecture built around a single elementary primitive: XOR-and-shift over GF(2). The architecture supports reversible variable binding (Pillar 1), non-commutative compositional bundling at fixed dimension (Pillar 2), and address-space individual/kind separation under the same algebra (Pillar 3). The treelet, in this reading, is an algebraic register set whose identity is a primitive generator polynomial; the stock of empty treelets is a library of distinct primitive polynomials per block; the filling-in of a treelet is the deposit of an Entry Address at the addressed block location. The biological substrate, as developed in the companion Perspective, is the dentate gyrus--CA3 circuit, with developmentally specified mossy-fiber targeting supplying exactly the kind of innate, non-blueprint microcircuitry the cascading-master-control-gene proposal anticipated.

We do not claim that the Algebraic Mind program is settled. The bit-exact reversibility of GF(2) operations holds in silicon but only approximately under biological noise; the correspondence between developmental programs and primitive-polynomial selection is structural rather than proven by functional-equivalence theorems; the treelet reinterpretation is one reading rather than the only possible reading; and no cognitive experiment is reported here. What we do claim is that the program has, for the first time, a worked-out substrate to compare against---an algebra under which symbol manipulation is operational, a substrate under which the algebra is biologically plausible, and a set of testable predictions that distinguish the engine reading from the alternatives Marcus considered in 2001.

Two consequences deserve to be carried out of this paper separately from the architecture that produced them.

The first is that compositionality is purchased with decomposability, and the purchase is made at the elementary operation or not at all. A binding that can only be approximately undone does not yield a slightly degraded constituent; it yields a different kind of object---a neighbourhood rather than a part---and no downstream cleverness recovers the difference. This is why Marcus's three pillars turn out not to be three requirements. They are one requirement inspected from three sides, and an architecture that fails it fails all three at once. We expect this constraint to hold well beyond the present substrate, and it cuts against the reflex of treating approximation as a quality dial that can be turned back up later.

The second is that the demand Marcus made of an inflection rule and the demand Pearl makes of a causal model are the same demand. Both ask what the system would do along a road it has not travelled; both are unanswerable by any representation that records outcomes rather than holding operations; and both are answered, when they are answered, by the ability to unbind at a site, substitute, and rebind without disturbing anything else. Free generalization and counterfactual reasoning are one capacity under two names, and the substrate that supplies either supplies both.

Both properties are, in the end, the same property seen from two sides: a representation that can be taken apart can be put back together differently, and a representation that cannot, cannot. That is the reversible, auditable symbol-manipulating layer we claim statistical minds structurally lack.

The Algebraic Mind, we suggest, has a concrete substrate.

\appendix

\section{Structural Flow Equivalence, Not Mathematical Isomorphism}

Throughout this paper, and more extensively in the companion Perspective, we assert correspondences between an algebra realised in silicon and mechanisms observed in tissue. It is worth setting out, in one place, what kind of claim that is, because the available vocabulary invites overstatement.

\emph{What is not claimed.} We do not claim mathematical isomorphism. An isomorphism demands a strict bidirectional structure-preserving map, and no such map is exhibited here or, to our knowledge, exhibitable on present evidence. Granule-cell ensembles are not elements of GF(2)$^L$; mossy-fiber targeting is not known to be equivalent to the selection of a primitive generator; dendritic nonlinearity is XOR-\emph{like} in its input--output signature and is not an XOR gate. Each of these gaps is real and none is closed by the argument of Section 8.

\emph{What is claimed.} We claim \emph{structural flow equivalence}: a correspondence in the computational flow ``division $\to$ local processing $\to$ integration,'' together with a correspondence in what each stage must accomplish for the whole to work. The dentate gyrus divides an input across many sparsely active units; each unit performs a local, structurally specified, nonlinear transformation; CA3 contracts the resulting local verdicts into one coherent reading. The engine divides a hypervector across blocks; each block performs a local, polynomial-specified, reversible diffusion; majority voting contracts the block outputs into one reading with confidence CR1. The two descriptions are compatible at the level of flow and of function, and incompatible at the level of substrate, and the claim covers the first two.

\emph{Why this is the right framing.} The relationship we are pointing at is the one biology calls convergent evolution. Two lineages---one carbon, one silicon---faced the same computational problem under different constraints and arrived at structurally similar solutions, not because either copied the other but because the problem admits few solutions. Read this way, the correspondence carries information without carrying a claim of identity: it tells us that the constraints are tight, which is exactly what one wants to know when asking whether a substrate could have been otherwise. Read as mimicry, the same correspondence would be a much weaker and much more easily refuted claim, since the engine plainly does not mimic anything---it was designed for speed and power on commodity silicon, and the convergence was noticed afterwards.

\emph{What would strengthen it.} The claim would be strengthened by any of the following: a demonstration that developmental specification of mossy-fiber targets selects transformations with the maximal-period property that primitive polynomials have; a measurement of population-level reversibility in DG--CA3 under sparse coding that meets a stated numerical bound; or a functional-equivalence theorem relating the contractive readout of CA3 attractor dynamics to block-wise majority voting under a specified noise model. Each is a research programme rather than an experiment, and we state them so that the present claim can be seen for the modest thing it is.

\paragraph{Acknowledgments.} We thank Gary Marcus for the framework that this paper builds on, and for the methodological clarity with which the 2001 program was stated. The open questions that motivated this work were articulated explicitly in \emph{The Algebraic Mind}; our contribution is to offer candidate answers.


\end{document}